\ifx\pfmtname\undefined
\documentclass[a4paper]{article}
\else
\documentclass{ujarticle}
\fi
\usepackage{robomech_en}
\usepackage{graphicx}

\begin{document}
\makeatletter
\title{YOLO and K-Means Based 3D Object Detection Method
\\on Image and Point Cloud}
{}
{}
{}

\author{ \small
\begin{tabular}{ll}
 $\bigcirc$ & Xuanyu YIN, Student Member, (AIST,The University of Tokyo), yinxuanyu@g.ecc.u-tokyo.ac.jp\\
  & Yoko SASAKI, Member, AIST, y-sasaki@aist.go.jp                                                                                                                                                                                                                                                                                                                                                                                                                                                                                                                                                                                                                                                                                                                                                                                                                                                                                                                                                                                                                                                                                                                                                                                                                                                                                                                                                                                                                                                                                                                                                                                                                                                                                                                                                                                                                                                                                                                                                                                                                                                                                                                                                                                                                                                                                                                                                                                                                                                                                                                                                                                                                                                                                                                                                                                                                                                                                                                                                                                                                                                                                                                                                                                                                                                                                                                                                                                                                                                                                                                                                                                                                               \\
  & Weimin WANG, Member, AIST, weimin.wang@aist.go.jp                                                                                                                                                                                                                                                                                                                                                                                                                                                                                                                                                                                                                                                                                                                                                                                                                                                                                                                                                                                                                                                                                                                                                                                                                                                                                                                                                                                                                                                                                                                                                                                                                                                                                                                                                                                                                                                                                                                                                                                                                                                                                                                                                                                                                                                                                                                                                                                                                                                                                                                                                                                                                                                                                                                                                                                                                                                                                                                                                                                                                                                                                                                                                                                                                                                                                                                                                                                                                                                                                                                                                                                                                                      \\
  & Kentaro SHIMIZU, Member, The University of Tokyo, shimizu@bi.a.u-tokyo.ac.jp
 \\
\end{tabular}
}
\makeatother

\abstract{ \small
Lidar based 3D object detection and classification tasks are essential for automated driving(AD). A Lidar sensor can provide the 3D point coud data reconstruction of the surrounding environment. But the detection in 3D point cloud still needs a strong algorithmic challenge. This paper consists of three parts.(1)Lidar-camera calib. (2)YOLO, based detection and PointCloud extraction, (3) k-means based point cloud segmentation. In our research, Camera can capture the image to make the Real-time 2D Object Detection by using YOLO,  I transfer the bounding box to node whose function is making 3d object detection on point cloud data from Lidar. By comparing whether 2D coordinate transferred from the 3D point is in the object bounding box or not, and doing a k-means clustering can achieve High-speed 3D object recognition function in GPU.
}

\date{} 

\maketitle
\thispagestyle{empty}
\pagestyle{empty}

\small
\section{Introduction}
Recently, great progress has been made on 2D image un-derstanding tasks, such as object detection and instancesegmentation [13]. However, after getting the 2D bounding boxes or pixel masks, the detection on 3D point cloud data becoming more and more necessary in many applications areas, such as Autonomous Driving(AD) and Augmented Reality(AR). More and more company or university laboratory come to use the 3D sensors to capture the 3D data. In this paper, I try to learn and make the experiment of the 3D Object Detection task which is one of the most important tasks in 3D computer vision and analysis the experiment result and talk about the future work which is needed to be done to imporve the 3D object detection average precision(AP).

In Autonomous Driving(AD) field, the LIDAR sensor is the most popular 3D sensors which will generate the 3D point clouds and capture the 3D structure of the scenes. The difficulty of point cloud-based 3D object detection mainly liesin  irregularity  of  the  point  clouds  from  LIDAR  sensors [2]. So State-of-art 3D object detection methods either leverage the mature 2D detection framework by projecting the point clouds into bird's view, to frontal view [2]. But in this progress, the information of point cloud will loss when we making the quantization process.

The aim of this paper is to extract the every point which may be the object after transforming in the 2D bouding box. Firstly, we make a device with 6 cameras and one LIDAR. Then we make the experiment to capture the 2D image by camera and 3D point cloud by LIDAR and store the data in a rosbag which can be reused in later experiment. Extracting of the image data which is distorted is necessary, but the undistorted transform progress is needed too. After the undistorted tranform progess. There will be 5 images which will be splited and dropped into the YOLO detection progress. The YOLO detection progress will return the realted bounding box and class label.I store the bouding box and the class label for later reading, reducing the coupling of project research. In the second step, I first wrote a function to extract the different topic information of rosbag. After extracting the point cloud file, I will put it into the numpy matrix for future operations.

After matching the 2D images corresponding to each point cloud by matching the fps of camera and LIDAR, data conversion based on external parameters and internal parameters is performed for each point cloud. For each different boudingbox of each point cloud, we collect all the matching points and render different colors based on different class labels. Finally, we make an unsupervised clustering of point clouds in different bounding boxes, remove some noise, and finally get the result of 3D object recognition. Save the recognition result into a rosbag, and then perform 3D visualization to check the experimental results. The above is the rough research process of this article.
\section{3D object detection method}

\subsection{Overview}
\begin{figure}[h]
\centering
\includegraphics[height=75mm]{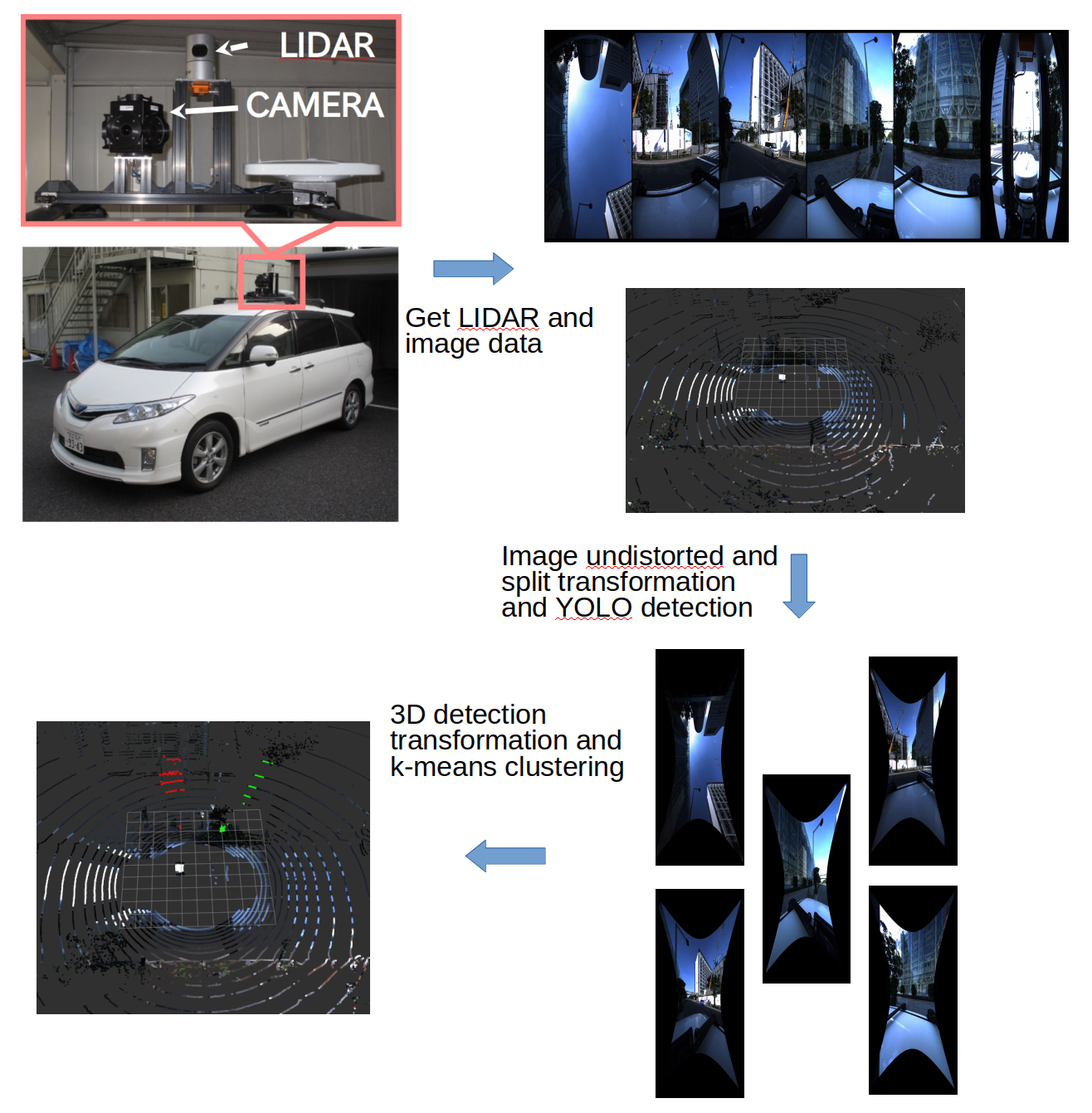}
\caption{Overview of Experiment step}
\end{figure}
This research is basically divided into five parts. The first part mainly focuses on the calibration of the camera and the structural design of the testing equipment. The second part is to convert the distorted image into an undistorted image. The third part is to use YOLOv3 to do object recognition on 2D images. This article mainly uses YOLOv1 tiny and YOLOv3 methods to do experiments, using keras to reproduce YOLO. The fourth part is the extraction of point clouds. This article uses rosbag to store data and RVIZ for point cloud visualization. The fifth part is the unsupervised clustering of kmeans, which is used to optimize the detection results of basic experiments and improve the detection accuracy of 3D object recognition.
\subsection{Lidar-camera calibration}
Here is mainly the information of the equipment used in this experiment and the external reference of the camera.

This experiment uses Velogyne lidar with five cameras to achieve 360 no dead angle monitoring which is shown in Figure 1.
\begin{figure}[h]
\centering

\includegraphics[height=80mm]{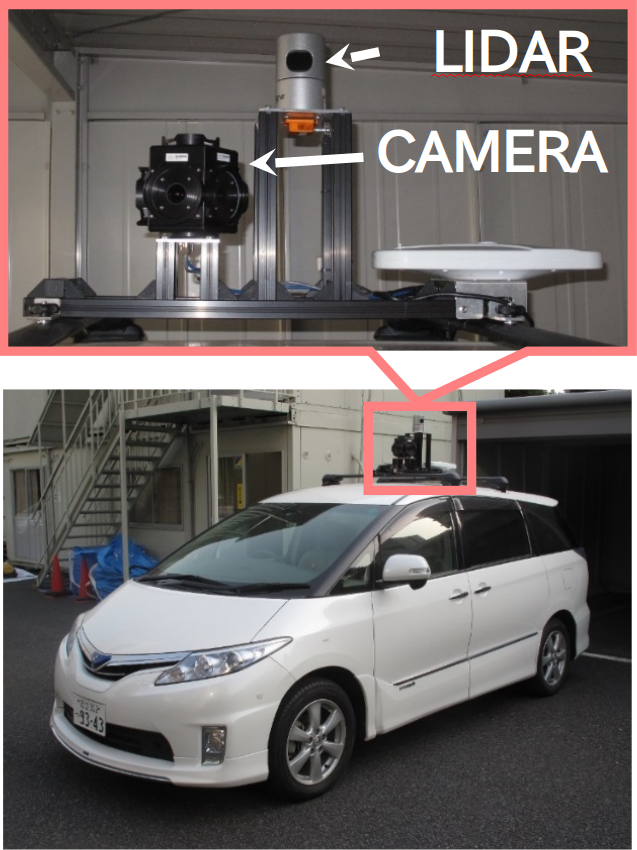}
\caption{Structure of detection device: LIDAR and omni-directional camera}
\end{figure}

In the image measurement process and machine vision application, in order to determine the relationship between the three-dimensional geometric position of a point on the surface of a space object and its corresponding point in the image, a geometric model of camera imaging must be established. These geometric model parameters are camera parameters. Under most conditions, these parameters must be obtained through experiments and calculations[17]. This process of solving parameters is called camera calibration (or camera calibration). The internal parameters of the five cameras obtained at the end of this paper are shown in the Fig.2.
\begin{figure}[h]
\centering
\includegraphics[height=25mm]{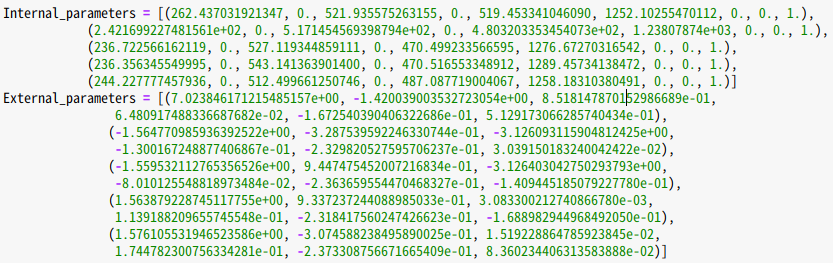}
\caption{Internal and external parameters}
\end{figure}
\subsection{Image undistorted transform}
In photography, it is generally believed that a wide-angle lens is prone to barrel distortion. Telephoto lenses are prone to pincushion distortion. If the camera uses a short focal length wide-angle lens, the resulting image will be more susceptible to barrel distortion because the magnification of the lens gradually decreases as the distance increases, causing the image pixels to radially surround the center point. Use opencv for image correction and camera calibration, the Fig.3 is the raw iamge and the Fig.4 is the image after the undistorted tranformation.
\begin{figure}[h]
\centering
\includegraphics[height=30mm]{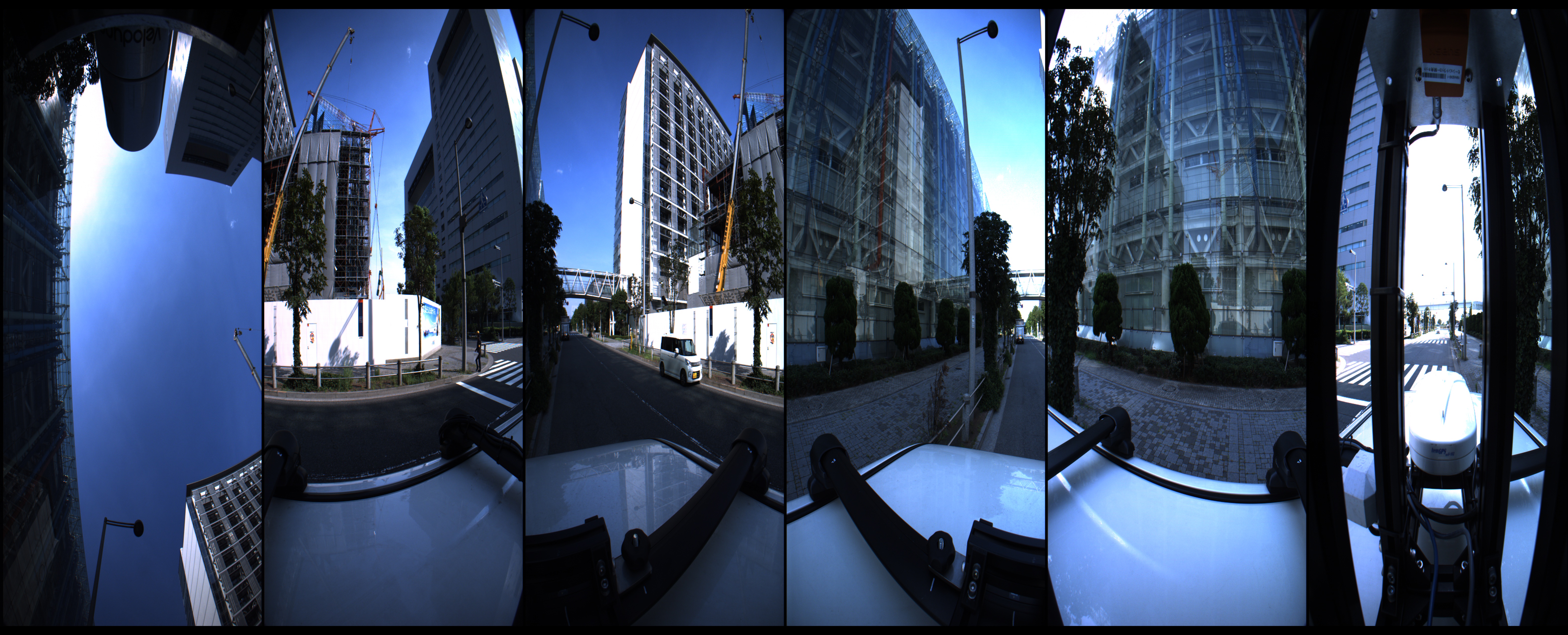}
\caption{Raw image of 5 cameras}
\end{figure}
\begin{figure}[h]
\centering
\includegraphics[height=30mm]{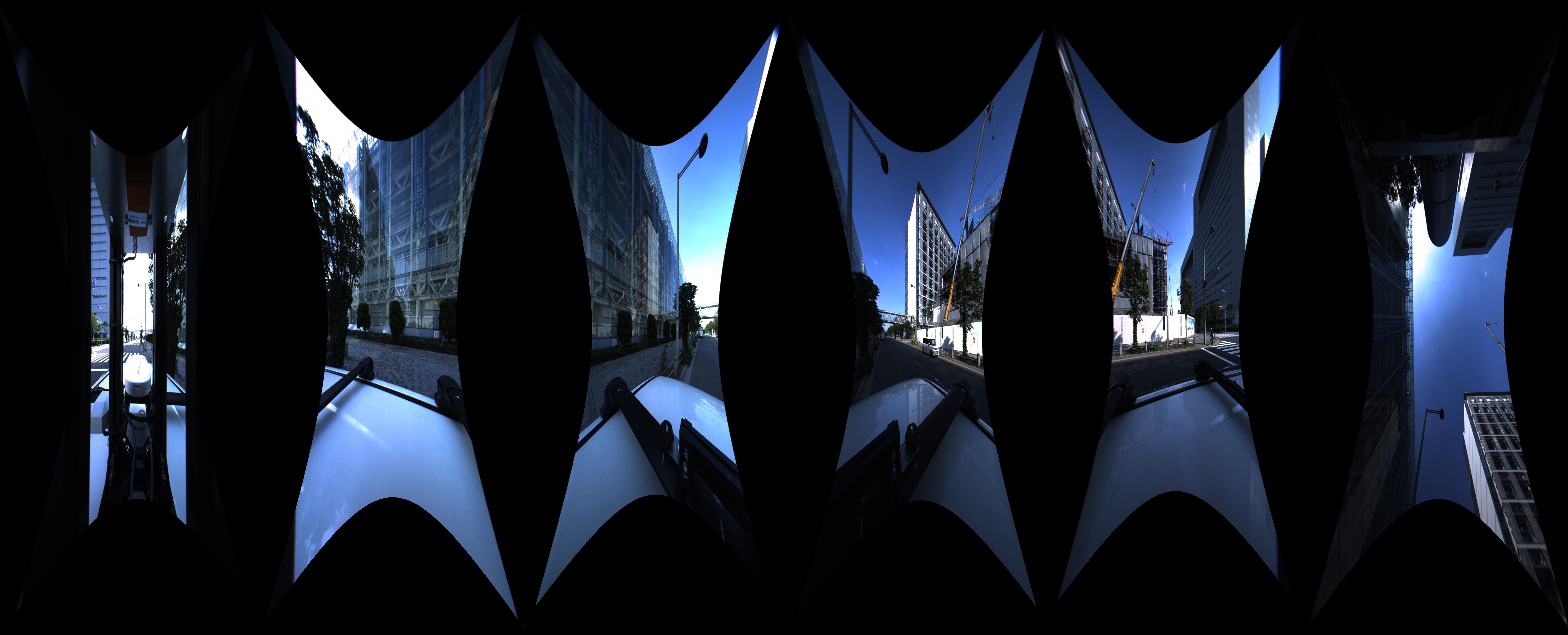}
\caption{Image of 5 cameras after undistorted transform}
\end{figure}
\subsection{YOLO based detection}
YOLO is a fast target detection algorithm that is very useful for tasks with very high real-time requirements. YOLO authors launched the YOLOv3 version in 2018. When training on Titan X, v3 is 3.8 times faster than RetinaNet in the case of mAP, and YOLOv3 can execute a 320×320 picture in 22ms, mAP. The score is 51.5, which is comparable to the accuracy of the SSD, but three times faster than it. YOLOv3 is very fast and accurate. In the case of IoU=0.5, it is equivalent to the mAP value of Focal Loss, but it is 4 times faster. So this article tried YOLOv3 as a 2D object detection algorithm, and the Fig.5 is the detection example from camera image, the Fig.6 is the 3D object detection example of point cloud.

This article has a total of all the classes that can be identified in the coco dataset, including person, bicycle, car, motorbike, aeroplane, bus, train, truck, boat, traffic, light, fire, hydrant, stop and so on, 80 classes in total.

However, most frequently classes must be truck, person and car. So maybe train a new YOLOv3 neural network which is aimed to detect these three classes will be useful in saving detection time and real time function.

Since the image is largely black after the undistorted conversion, you must remove the noise that exceeds its maximum boudingbox when doing the conversion. Because the boundingxbox of more than a quarter of the image size will contain black parts, this part must be noisy.
\begin{figure}[h]
\centering
\includegraphics[height=40mm]{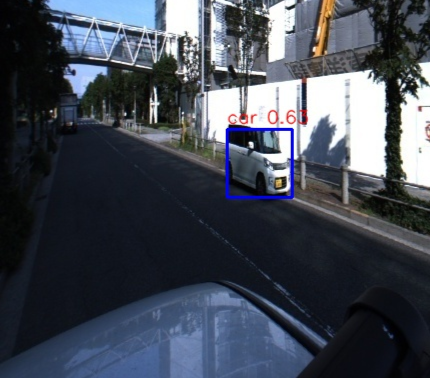}
\caption{YOLO detection example from camera image}
\end{figure}
\begin{figure}[h]
\centering
\includegraphics[height=45mm]{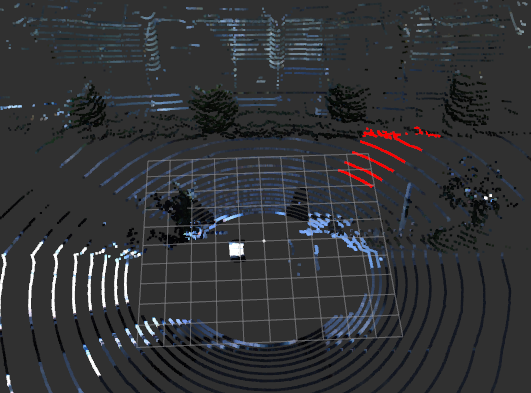}
\caption{3D object detection example of point cloud}
\end{figure}
\subsection{PointCloud extraction}
This article mainly uses rosbag to read and read data. The read data contains undistorted image and point cloud image. The output is mainly the result of point cloud detection.
\subsection{K-means based point cloud segmentation}
\begin{figure}[h]
\centering
\includegraphics[height=35mm]{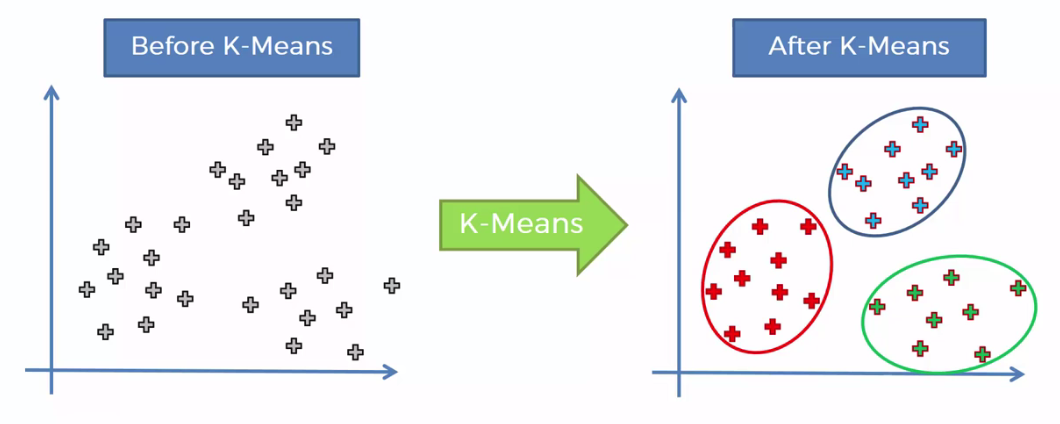}
\caption{K-means clustering}
\end{figure}

Since 2D is converted into 3D, all points that can be mapped into the boudingbox in a certain direction will be marked with different label colors. In order to make the experiment better, we have tried the kmeans machine learning unsupervised clustering method. The detection is faster, and the accuracy of the points is much improved, removing most of the noise points, the K-means clustering graph is shown in Fig.8.

\section{Experiments and Results}
The in-vehicle sensor is used here, a large amount of data is collected, the identification experiment is the part of the data used, and the experimental results are finally evaluated.
\subsection{Experiments with different classes}
This paper first made a direct conversion 3D prediction result. Basically, all the points on the image that can be mapped to the boundingbox will be recognized. This leads to a lot of noise, and it is impossible to prepare to identify the 3D boundingbox, but can recognize the 3D radar data. The specific category exists in which direction, thus completing the rough 3D recognition function. The 2D YOLO example is shown in Fig.10, and the related experiment result with and without k-means are shown in Fig.11 and Fig.12.

The proofreading here is mainly for the naked eye. According to the specific boundingbox data and the data under the camera, then judge whether the recognition result is correct or not.
\begin{figure}[h]
\centering
\includegraphics[height=40mm]{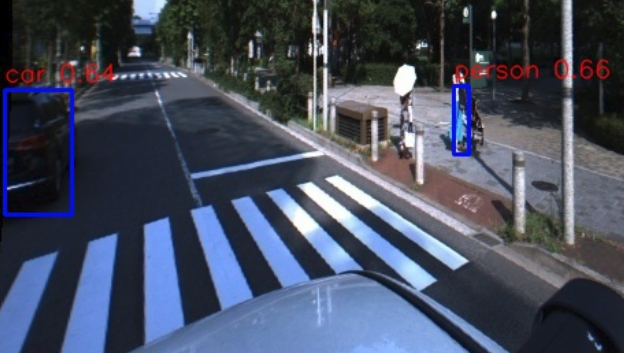}
\caption{2D YOLO experiment result}
\end{figure}
\begin{figure}[h]
\centering
\includegraphics[height=45mm]{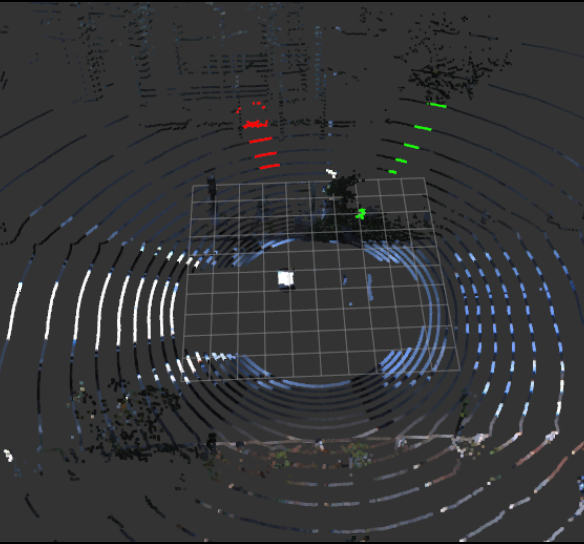}
\caption{3D experiment result without k-means}
\end{figure}
\subsection{Experiments with K-Means}
k-means clustering is a method of vector quantization, originally from signal processing, that is popular for cluster analysis in data mining. k-means clustering aims to partition n observations into k clusters in which each observation belongs to the cluster with the nearest mean, serving as a prototype of the cluster. This results in a partitioning of the data space into Voronoi cells [15].

Since kmeans pre-determines the number of categories and the maximum number of cluster iterations, the center point is randomly taken at first, so the result of each cluster will be biased, but the expected value of each category will not change too much at the end of the experiment. 

After using YOLO to complete the 2D object recognition and convert the data into 3D point cloud data, this paper mainly uses kmeans to further cluster the points of the corresponding bounding box that have been acquired, so as to remove some noise and make the recognition result more accurate.
\begin{figure}[h]
\centering
\includegraphics[height=43mm]{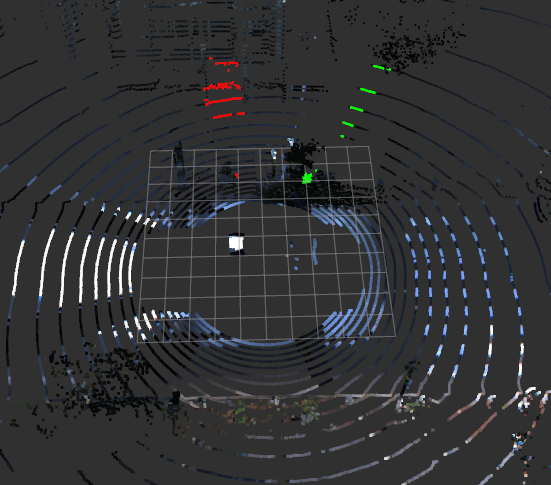}
\caption{3D experiment result with kmeans}
\end{figure}
\subsection{Result with and without unsupervised learning}
The results of the experiment with and without kmeans refer to Figure 13 and Figure 14. The total number of point clouds in each group is 232,320. The maximum number of points is 9215 points, This is when a car or truck is very close. The least number is 0, which means that there is no target object that can be recognized around at that time. In other words, it is an empty space with no pre-trained class label objects.

Figure 13 represents the ratio of dropped data after clustering using the kmeans method on the point cloud dataset. The highest is at 49.43 percent at 46 frames. In the case of not recognizing that no object is recognized, the lowest is at the 26 frames, only 5.53 percent.

When clustering is not performed, the experimental results will stain all the points mapped to the boudingbox in that direction, and after clustering, because if there is an object, the number of point clouds in that part will be significantly increased, while other places have no The object will obviously notice a decrease in the number of clouds. Therefore, unsupervised clustering can significantly change the results, and generally removes 30 percent of the noise.

\begin{figure}[h]
\centering
\includegraphics[height=45mm]{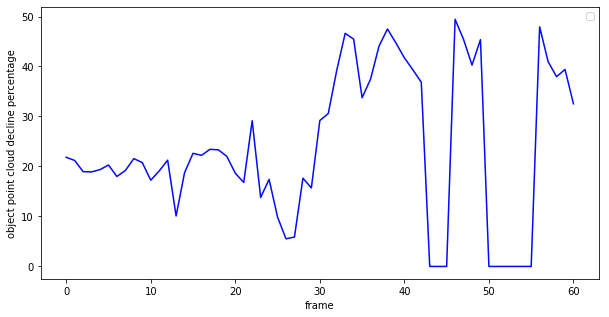}
\caption{Rate of point cloud decline with k-means function}
\end{figure}

\begin{figure}[h]
\centering
\includegraphics[height=45mm]{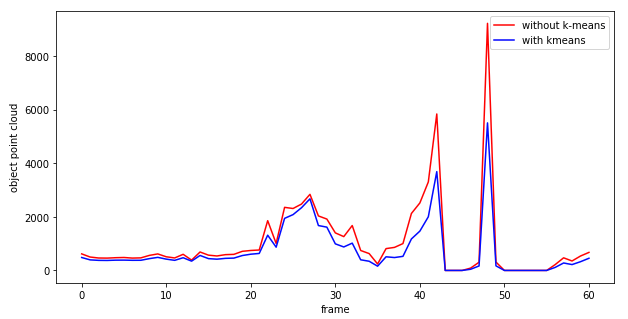}
\caption{Experiment result of detection point cloud number with and without k-means}
\end{figure}
\section{Conclusions}
Charles et al. at Stanford University published a paper on CVPR in 2017, proposing a deep learning network that directly handles point clouds, pointnet. This paper has a milestone significance, marking the point cloud processing has entered a new stage, because before the pointnet, the point cloud has no way to deal with it directly. Because the point cloud is three-dimensional, no smooth, not to mention the deep neural network, that makes many ordinary algorithms can not work, so many people have come up with a variety of methods, such as flattening the point cloud into a picture (MVCNN) , divide the point cloud into voxels, and then divide the point cloud into nodes and straighten them in order. When pointnet appeared, from this point the cloud domain became "pre-pointnet era" and "post-pointnet era". After PointNet, PointCNN, SO-NET and so on comes out, the operation of these methods are getting better and better.

The method adopted by this paper is to directly convert the 3D point cloud to 2D image data, from the recognition of the 2D boudingbox to the dyeing of the 3D point cloud. Since the YOLO algorithm is adopted, the real-time performance is very strong, and the unsupervised clustering is used too. A lot of noise will be removed. It makes the recognition better.

This paper mainly wants to find a way to quickly and accurately determine whether there are objects and objects in a certain direction. This will contribute to the success of the unmanned field, allowing the car to obtain more information to make more judgments.

The final experimental results, in the case of using two 1080Ti GPUs, basically ensure that the experiment without clustering consumes 0.95 seconds per frame and 0.96 seconds after k-means clustering. The fast identification process ensures the real-time detection of the surrounding conditions in unmanned driving. If multi-threading, parallel, distributed computing and other technologies are used, the recognition speed will be faster.

\section{Future work}
In the future, the position of robot will be combined and the semantic mapping from running mobile robot will be the core of next step. Not only the K-means function, but also the PointNet and FCN and so on will be considered in next step in proposed method. The By using the proposed method, automatic labeling function will be developed for training data generation of LIDAR based 3D object detection.

\end{document}